\documentclass[lettersize,journal]{IEEEtran}
\usepackage{amsmath,amsfonts}
\usepackage{algorithmic}
\usepackage{algorithm}
\usepackage{array}
\usepackage[caption=false,font=normalsize,labelfont=sf,textfont=sf]{subfig}
\usepackage{caption}
\usepackage{textcomp}
\usepackage{stfloats}
\usepackage{url}
\usepackage{verbatim}
\usepackage{indentfirst,bm}
\usepackage{graphicx}
\usepackage{pifont}
\usepackage{multirow}
\usepackage{epstopdf}
\usepackage{amsthm}
\usepackage{color}

\def\BibTeX{{\rm B\kern-.05em{\sc i\kern-.025em b}\kern-.08em
    T\kern-.1667em\lower.7ex\hbox{E}\kern-.125emX}}
\usepackage{balance}
\begin{document}
\title{Pixel-Wise Symbol Spotting via Progressive Points Location for Parsing CAD Images}

\author{Junbiao Pang, Zailin Dong, Jiaxin Deng, Mengyuan Zhu and Yunwei Zhang
\IEEEcompsocitemizethanks
{
\IEEEcompsocthanksitem J. Pang, Z. Dong and J. Deng are with the Faculty of Information Technology, Beijing University of Technology, Beijing 100124, China (e-mail: \mbox{junbiao\_pang@bjut.edu.cn}).

\IEEEcompsocthanksitem  M. Zhu and Y. Zhang are with the Software Development Center, China Information Technology Designing and Consulting Institute Co., Ltd, Beijing
100190, China.
 }
}


\maketitle

\begin{abstract}
Parsing Computer-Aided Design (CAD) drawings is a fundamental step for CAD revision, semantic-based management, and the generation of 3D prototypes in both the architecture and engineering industries. Labeling symbols from a CAD drawing is a challenging yet notorious task from a practical point of view. In this work, we propose to label and spot symbols from CAD images that are converted from CAD drawings. The advantage of spotting symbols from CAD images lies in the low requirement of labelers and the low-cost annotation. However, pixel-wise spotting symbols from CAD images is challenging work. We propose a pixel-wise point location via Progressive Gaussian Kernels (PGK) to balance between training efficiency and location accuracy. Besides, we introduce a local offset to the heatmap-based point location method. Based on the keypoints detection, we propose a symbol grouping method to redraw the rectangle symbols in CAD images. We have released a dataset containing CAD images of equipment rooms from telecommunication industrial CAD drawings. Extensive experiments on this real-world dataset show that the proposed method has good generalization ability.
\end{abstract}

\begin{IEEEkeywords}
 CAD images parsing,symbol spotting, pixel-wise localization, Progressive Gaussian Kernels
\end{IEEEkeywords}

\section{Introduction}

\IEEEPARstart{C}{omputer-Aided Design}(CAD) drawings are widely employed for efficient yet precise illustration of products, aiding in the creation, modification, analysis, management, and optimization processes throughout their entire life cycle~\cite{zheng2022gat}~\cite{fan2022cadtransformer}~\cite{liu2024symbol}. Therefore, CAD drawings are extensively utilized in the modern architecture, engineering, and construction (AEC) industries. Currently, many CAD drawings are still stored in the paper format. Semantic understanding of these technical documents in digital libraries becomes necessary due to the following practical requirements:
\begin{itemize}
\item \emph{Standardization}: It is necessary to ensure that symbols representing the same objects across different design units or individual engineers adhere to the same drawing standards for the effective communication and reusability.
\item \emph{Management}: For an asset owner, indexing, summarizing, and quantifying these assets via CAD drawings would efficiently manage their properties. It is challenging to automatically extract information, such as the number of spare frameworks. In fact, a substantial number of drawings in paper or image form result in several drawbacks during actual data storage, retrieval, and editing~\cite{li2022free2cad}.
\item \emph{Re-usability}: Designers or engineers usually reuse CAD drawings from the previous projects for their new designs. In this scenario, an efficient solution is to query semantic symbols from CAD drawings~\cite{zheng2022gat}~\cite{fan2021floorplancad}.
\end{itemize}


\begin{figure}[t!]
    \centerline
    {\includegraphics[width = 0.5\textwidth]{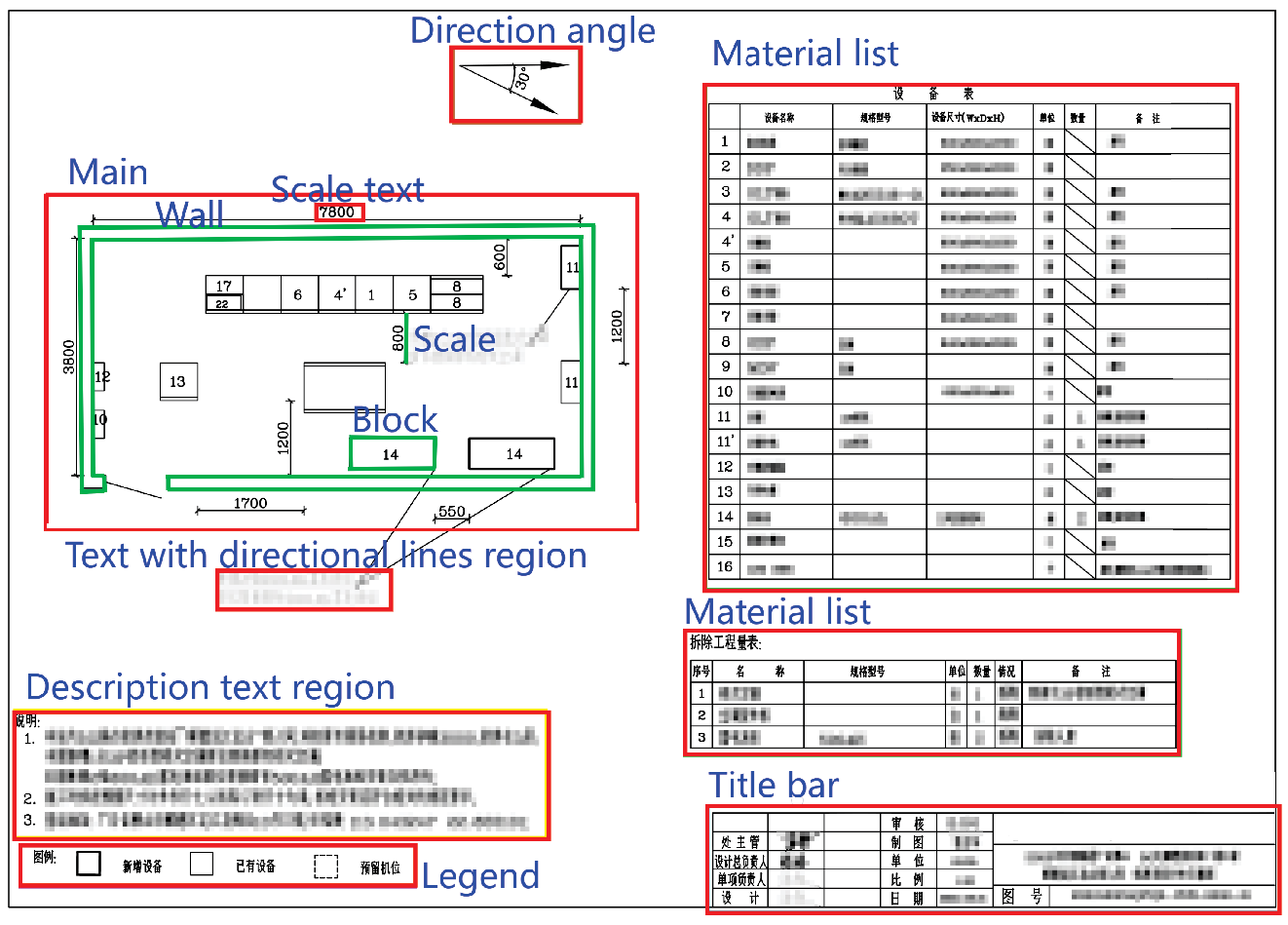}}
    \caption{12 kinds of semantic objects are defined within CAD images. A red box represents the region symbol, while the green wireframes represent the rectangle symbols.}
    \label{fig:group-meaningful-entities}
\end{figure}

Symbol spotting and parsing~\cite{rezvanifar2021analyzing} refer to a particular application of computer vision, in which symbols with the domain-specific semantics are localized and recognized to the predefined types~\cite{bhanbhro2023symbol}~\cite{rezvanifar2020symbol}. Currently, symbol spotting includes both the domain-specific locations of object instances (\textit{i.e.}, windows, doors, walls and body frames) and the identification of their attributes (\textit{i.e.}, the open direction of a door, the width of a wall and the size of a window). Spotting symbols from the CAD drawings is crucial to the above real-world industrial applications. Traditional symbol spotting~\cite{rezvanifar2019symbol}~\cite{sarkar2022automatic}for CAD drawings detects instance by classifying the elements (\textit{e.g.}, points, lines, arcs) into different symbols. However, the annotation of CAD drawings requires that annotators should skillfully use a CAD software to group graphical primitives (\textit{e.g.}, arrows, lines, circles) into semantic symbols, as shown in Fig.\ref{fig2-compare-cad-image-and-annotation}(a). The efficient yet low-cost annotation of CAD drawings is a pressing and practical problem that hinders symbol spotting for CAD drawings.

In this paper, we propose to spot symbols from CAD images that are converted from CAD drawings. The motivations for spotting symbols from CAD images are threefold: 1) the requirements for labelers are lower compared to those for CAD images; 2) CAD drawings in papers or images are handled in a uniform approach; 3) CAD images only require labelers to annotate the keypoints or lines of symbols, as shown in Fig.\ref{fig2-compare-cad-image-and-annotation}(b).

Technically, the symbol spotting for CAD images can be further classified into two sub-tasks as follows:
\begin{itemize}
\item \emph{Object-Based Spotting (OBS)}: OBS utilizes off-the-shelf methods to coarsely locate symbols, severing as layout analysis~\cite{wu2021document}~\cite{binmakhashen2019document} .
OBS usually unifies symbols that have the same semantics but different visual representations into uniform symbols. Empirically, the object detection method ( \textit{e.g.}, YOLOX~\cite{ge2021yolox})achieves excellent performance for the object-level sub-task.

\item \emph{Pixel-Based Spotting (PBS)}: PBS precisely decomposes a symbol into a set of keypoints (\emph{e.g.}, the start point and the end one of a line), aiming to locate and redraw this symbol in a pixel-wise approach. PBS paves the way for reusing, changing, or adding symbols for management requirements.A significant challenge of PBS is the pixel-wise keypoint location since most of the images converted from CAD drawings are high resolution, such as, $4000\times 4000$. Therefore, PBS is a non-trivial task for parsing CAD images.
\end{itemize}

The state-of-the-art(SOTA) keypoint detection approaches use the Gaussian heatmap to represent a point~\cite{toshev2014deeppose}~\cite{xiao2018simple}~\cite{sun2019deep}~\cite{yu2021lite} for Human Pose Estimation (HPE). However, HPE is totally different to the keypoint location in CAD images due to the following aspects: 1) keypoints in HPE themselves contain labeling uncertainty,  while keypoints in parsing CAD images barely have uncertainty; 2) HPE considers the scale of different objects, while parsing CAD images do not have the scale problem;3) Additionally, the appearances of keypoints in CAD images differ significantly from those in HPE~\cite{newell2016stacked}~\cite{qu2022heatmap} or object counting~\cite{liu2021counting}. Therefore,  pixel-wise keypoint location is a novel yet challenging task for parsing CAD images.

Naturally, for the PBS for CAD images, we seek an accurate and stable pixel-wise keypoint location method, based on two motivations. First, although an enormous volume of literature has been devoted to the accuracy of the point detection for HPE~\cite{wang2022lite}~\cite{dong2022cswin}~\cite{wang2020deep}, there is little attention to the pixel-wise point location method, and practitioners lack guidelines on how to achieve it. Second, we want to combine the advantages of the popular heatmap based method~\cite{nibali2018numerical}~\cite{li2022simcc}and the regression-based method~\cite{li2021human}~\cite{gu2021removing}~\cite{sun2018integral}: the heatmap-based method anchors the accuracy,  while the regression-based method enhances the prediction performance. In summary, we desire a pixel-wise keypoint prediction ability for the PBS for CAD images.

In this paper, we handle the pixel-wise keypoint location from the viewpoint of role of Gaussian Kernel Size (GKS) and the compensation of the quantization error. Concretely, a small GKS facilitates more accurate point location than a large GKS by finding the maximum response~\cite{gu2021dive}. However, a small GKS provides insufficient gradient information to train a neural network (as will be discussed in Fig.\ref{fig-gradient}). Conversely, a large GKS expedite the training of a neural network but is susceptible to be miss-guided by the Max-Value Drift (MVD) problem (as will be discussed in Fig.\ref{fig:combined_heatmap}) during the decoding process~\cite{sun2019deep}, rendering the use of large GKS unnecessary. However, a naive approach, switching the GKS during training,  has been empirically shown to result in an unstable training process and may even lead to non-convergence.

Motivated by the above analysis, we propose Progressive Gaussian Kernels (PGK) to harmonically use both a large GKS and a small one in an annealing approach without suffering from the complex position decoding process ~\cite{gu2021dive} for symbol spotting in CAD images. There are many potential benefits of the combination of the large GKS and small one: accelerating neural network training, and improved keypoint location accuracy. Besides, we introduce an offset in the position encoding stage to reduce the quantization error. Based on keypoint localization, we propose a symbol grouping method for the rectangle-like symbols. We have released an image dataset depicting the layout of equipment rooms from telecommunication industrial CAD drawings. We verify the effectiveness of our method on the proposed dataset.

To the best of our knowledge, this paper is the first to combine the large GKS and the small one for keypoint detection, presenting a comprehensive series of experiments to illustrate the benefits of this novel technique. The proposed method is computationally simple yet exceptionally powerful. Simply by annealing the heatmap size, with no further parameter tuning, we find an accurate keypoint location method that meets or exceeds the current state of the art!

\section{Related Work}

\subsection{Background of Symbol Spotting}

\begin{figure}[t!]
\begin{center}
    \centering
    \subfloat[A CAD drawing and its annotation]
    {
        \label{cad-drawing-and-annotation}\includegraphics[width = 0.5\textwidth]{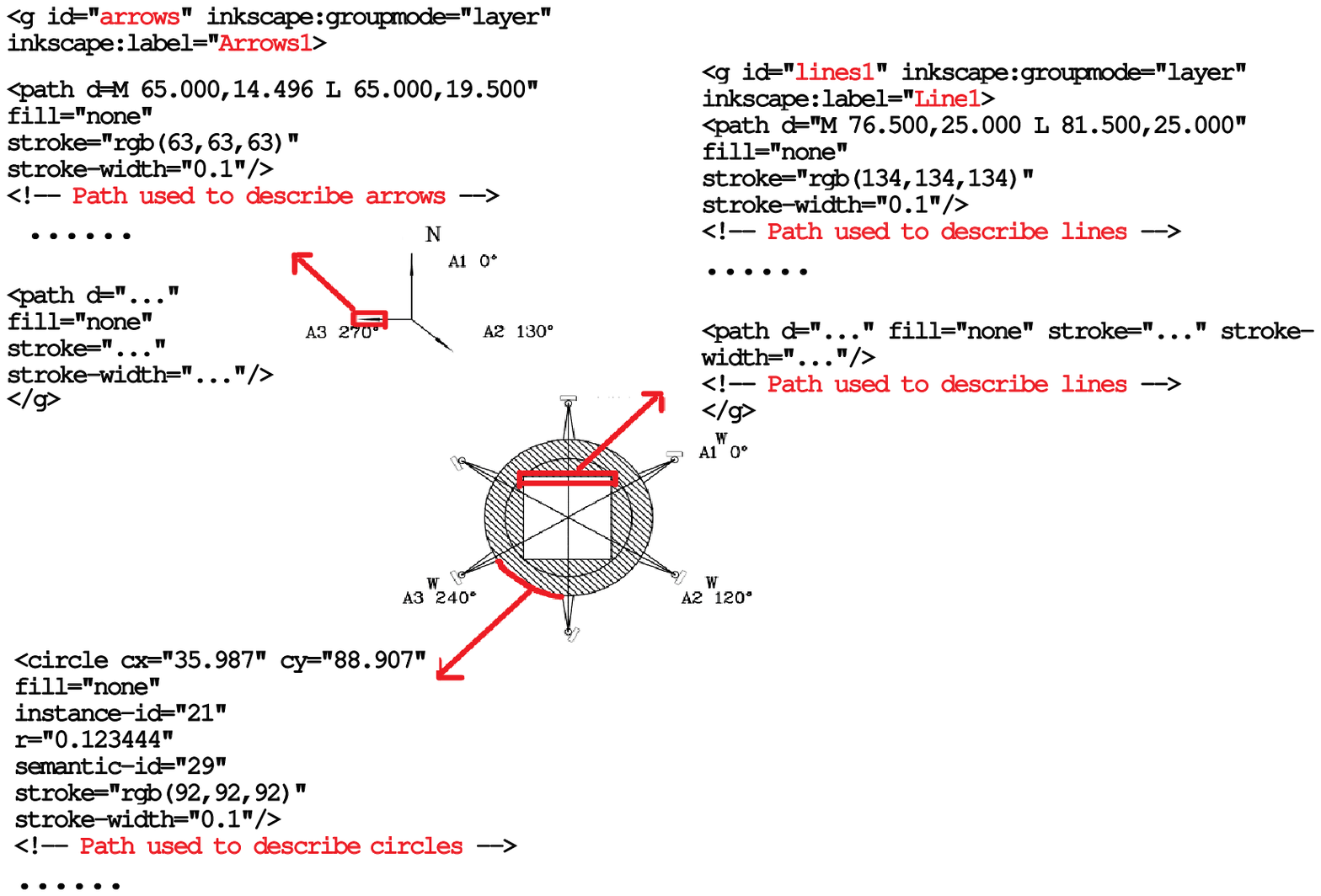}
    }
    \
    \subfloat[A CAD image and its annotation]
    {
        \label{cad-images-and-annotation}\includegraphics[scale=0.33]{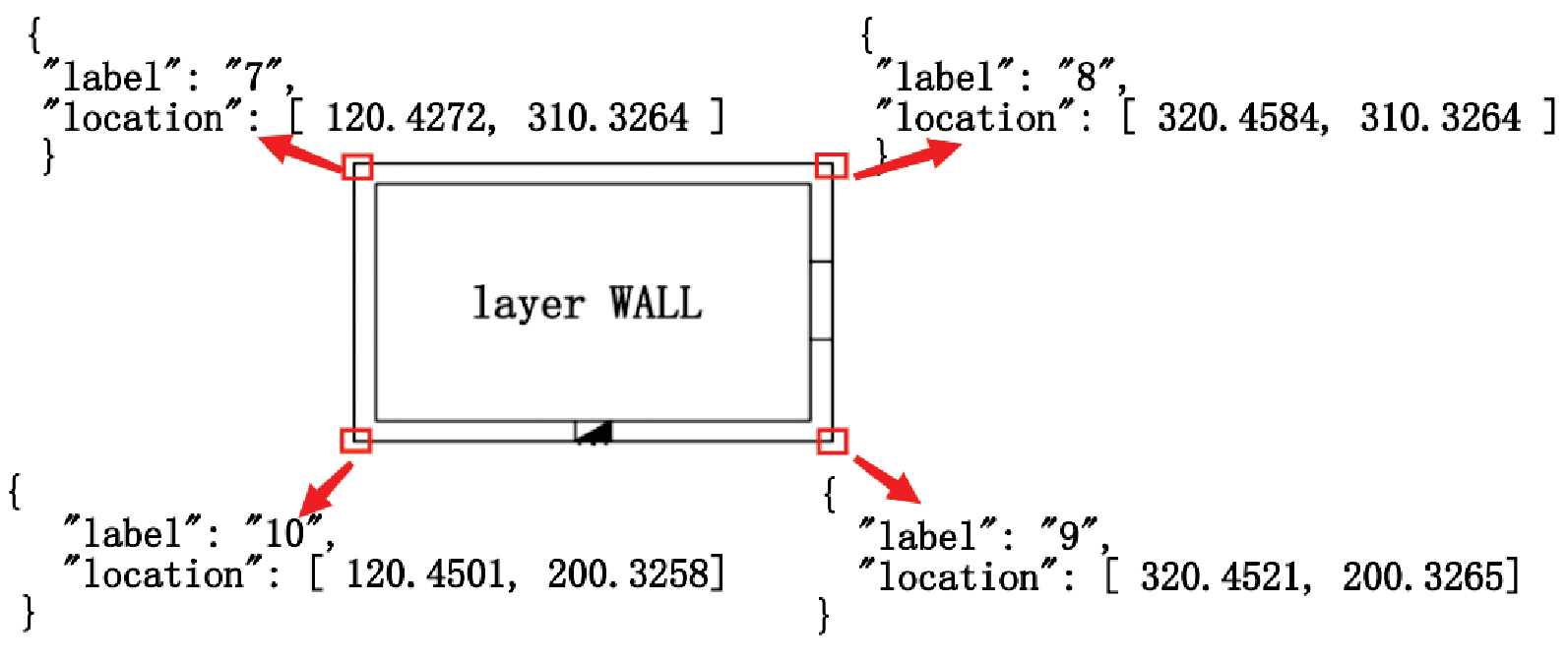}
    }
    \caption{Comparison between a CAD drawing and a CAD image.}
    \label{fig2-compare-cad-image-and-annotation}
\end{center}
\end{figure}

In practice, there are two research lines for symbol spotting: 1) the CAD drawings-based approach uses CAD format( \textit{e.g.}, .dwg)\cite{zheng2022gat}\cite{fan2022cadtransformer}~\cite{jianwu2024artificial}~\cite{you2022art} and 2) the image-based approach uses images (\textit{e.g.}, .png,. jpeg) converted from the CAD format files~\cite{fan2021floorplancad}~\cite{lu2020semi}. The differences between the two lines are summarized as follows:
\begin{itemize}
\item \emph{Pixel-wise location}: As illustrated in Fig\ref{fig2-compare-cad-image-and-annotation}, the CAD drawings supply precisely location of symbols in a structured form, while the image-based approach needs pixel-wisely to locate symbols from images which have the same size as CAD drawings, such as $4200 \times 3000$.
\item \emph{Efficient annotation}: The CAD drawings require annotators to skillfully use a CAD software to group graphical primitives (\emph{i.e.}, arrows, lines, circles and their segmentations) into semantic symbols (see Fig.~\ref{cad-drawing-and-annotation}). In contrast, the image-based approach only requires labelers annotate the keypoints or the lines of a symbol.
\item \emph{Feature representation}: The CAD drawing-based approach utilizes vector graphics extracted from CAD drawings as input, such as, clockwise angles, lengths and types. In contrast, the image-based approach relies solely on unstructured images themselves.
\item \emph{Data deficiency}: Nowadays many documents are still stored in paper format without the corresponding CAD drawing files. Consequently, the image-based approach naturally handles the deficiency of the CAD drawings.
\end{itemize}

\subsection{Symbol Spotting}

Conventional symbol spotting methods are categorized as the vector-based methods~\cite{rezvanifar2019symbol} and the image-based methods~\cite{detone2018superpoint}\cite{aminikhanghahi2017survey}\cite{ko2021key}. Concretely, the vector-based methods utilize the structural relation among symbol primitives, while the image-based methods leverage the statistical properties of pixels. The challenge is that the handcrafted symbol descriptors barely cope with the graphical notation variability of all kinds ~\cite{rezvanifar2020symbol}.

Deep learning has been adapted to the symbol spotting in the vector-based methods for CAD drawings, where the popular detection models Faster-RCNN~\cite{ren2015faster}, YOLO~\cite{jiang2022review}, Graph Neural Network (GNN)~\cite{zhou2020graph}, and Transformer~\cite{vaswani2017attention} have been exploited. However, these methods require the expensive primitives-level annotations, making it difficult to generalize to any vector-based documents. In practice, a symbol maybe drawn from different layers in CAD drawings, requiring professionals with CAD software to label primitives in a time-consuming approach. On the contrary, the image-based method only requires the pixel-wise keypoint annotation, reducing the requirements of labelers.

The image-based methods often locate the bounding boxes of symbols in CAD images via the  off-the-shelf object detectors~\cite{ren2015faster}~\cite{he2017mask}~\cite{schneider2016example}. However, previous image-based methods focus on the types of symbols rather than pixel-based symbol spotting from CAD images.

Recently, a real-world floor plan CAD drawing dataset~\cite{fan2021floorplancad} has been released, contains $35$ object classes of interest, including $30$ countable symbols and $5$ uncountable classes labeled with line-grained annotations in a vector-based document.
As a comparison, our dataset contains $15$ types of keypoints which are further organized into $12$ semantic symbols for CAD images, where the image resolutions range from $1700 \times1200$ to $4200\times3000$. To our best knowledge, we are the first to release the largest CAD images for symbol spotting via keypoint location.

\subsection{Keypoint Location}

Keypoint location methods are generally classified into two categories: the heatmap-based approach and the  regression-based one. The heatmap-based approach~\cite{tompson2015efficient}~\cite{andriluka20142d} fully utilized the spatial information around keypoints, thus achieving the higher accuracy than that of the regression-based one. Some studies~\cite{newell2016stacked}~\cite{wang2020deep}~\cite{yu2021lite} have proposed deep neural networks for feature extraction.
In contrast, the regression-based method directly outputs the coordinates of keypoints for HPE.  For example, CenterNet~\cite{duan2019centernet} and DirectPose~\cite{tian2019directpose} estimate multi-person HPE in a one-stage object detection framework. Recently, Residual Log-likelihood Estimation (RLE)~\cite{li2021human} introduced a normalized flow model~\cite{dinh2016density} to capture the underlying distribution of keypoints. Compared with the heatmap-based method, the regression-based approach has made great efforts to model the implicit relationships among keypoints, yet still lacks spatial generalization ability ~\cite{nibali2018numerical}. Experiments empirically show that the heatmap-based methods still obtain a better results than the regression-based ones~\cite{li2022simcc}.

It is important to note that the above keypoint location methods were primarily proposed for HPE, in which datasets include MPII~\cite{andriluka20142d} and COCO~\cite{lin2014microsoft}. HPE considers the possible occlusion, object scale and complex backgrounds where the evaluation metric, Object Keypoint
Similarity (OKS), is used to handle the uncertainty of keypoints. In contrast, symbol spotting for the CAD images requires the pixel-wise keypoint location. To obtain a pixel-base location for the CAD images, we propose to combine the heatmap-based method and regression-based one, utilizing local offsets to compensate for the quantization errors of keypoints.

\section{Progressive KeyPoints Detection}\label{section3}

\subsection{Heatmap based Coordinate Encoding-Decoding Revisited}

\begin{figure}[t!]
    \centering
    \subfloat[GT heatmap ($\sigma = 3$)]  
    {
        \includegraphics[scale=0.40]{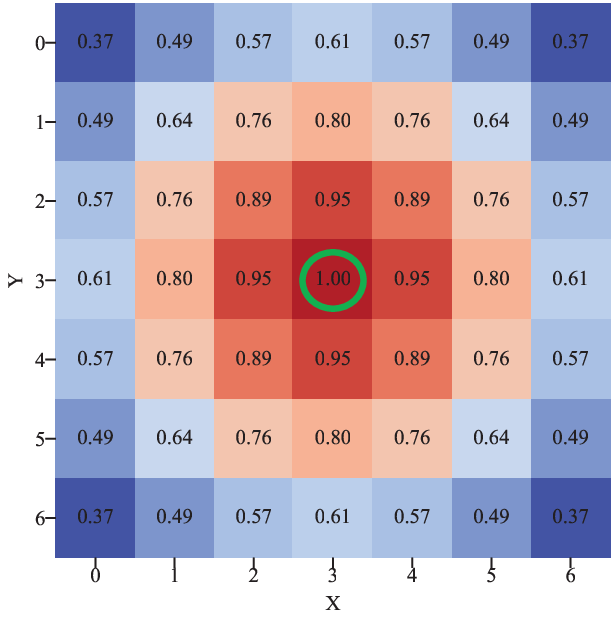}
        \label{fig:heatmap_sigma_3_groundtruth}
    }
    \subfloat[GT heatmap ($\sigma = 1$)]  
    {
        \includegraphics[scale=0.40]{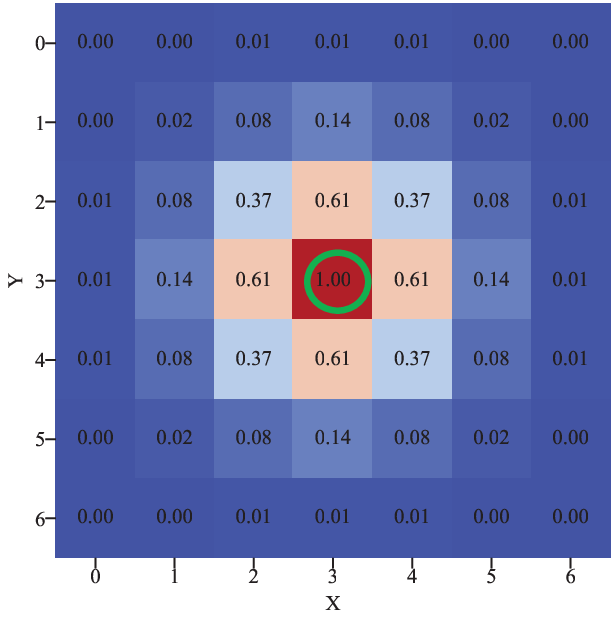}
        \label{fig:heatmap_sigma_1_groundtruth}
    }
    \\
    \subfloat[Predicted heatmap
    ($\sigma = 3$)]  
    {
        \includegraphics[scale=0.40]{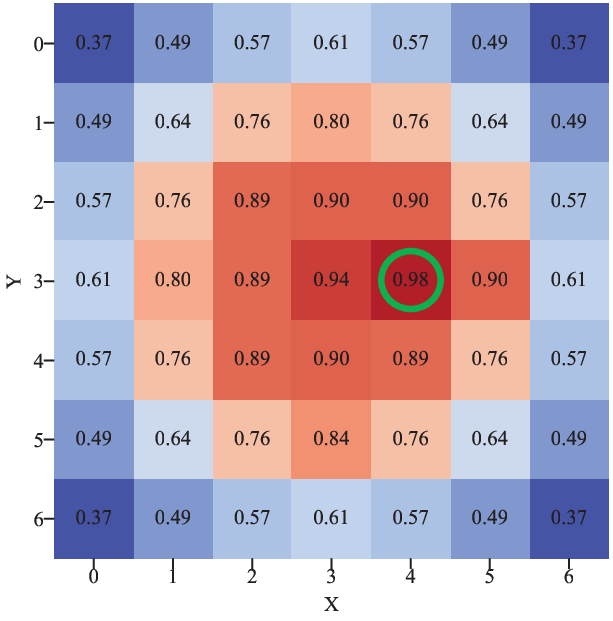}
        \label{fig:predictive_sigma_3}
    }
    \subfloat[Predicted heatmap
    ($\sigma = 1$)]  
    {
        \includegraphics[scale=0.40]{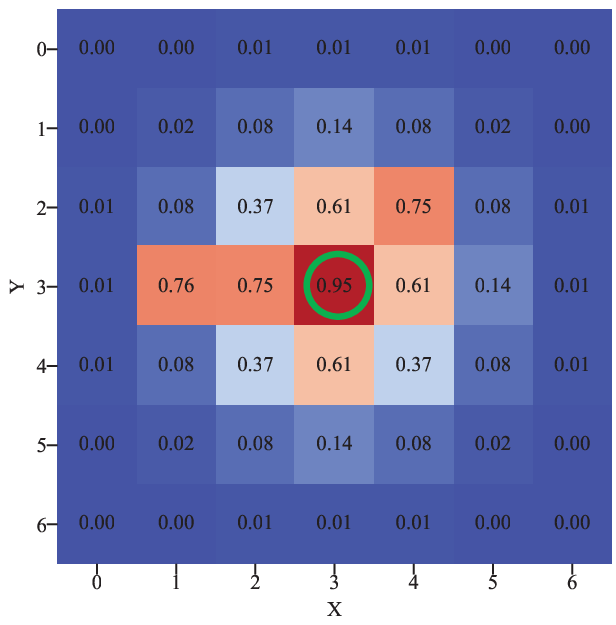}
        \label{fig:predictive_sigma_1}
    }
    \caption{Illustration of MVD problem by comprising between the GT heatmaps and the predicted heatmaps. (a) and (b) represent the GT heatmaps with the GKS $\sigma=3$ and $\sigma=1$,respectively. (c) and (d) show the predicted heatmaps with the corresponding GKS respectively. Color represents the magnitude of the heatmaps response. Circle indicates the position with the maximum value.}
    \label{fig:combined_heatmap}
\end{figure}

The encoding of a keypoint is the process of converting the a coordinate into a Gaussian distribution. Specifically, given a heatmap  $H\in \mathbb{R}^{h \times w}$ with the height $h$ and the width $w$, we assume the Ground Truth (GT) of a keypoint to be $\bm{{\mu}} =[\mu_x,\mu_y]$. The heatmap of a GT keypoint is converted into a Gaussian distribution as follows:
\begin{equation}
\label{eqt1:gaussionGT}
H\left( x,y \right) = \frac{1}{\sqrt {2\pi } {\sigma }} \exp \left( -\frac{(x - \mu_x)^2 + (y - \mu_y)^2}{2\sigma^2} \right)
\end{equation}
where ($x$,$y$) is the discrete pixel coordinate in the heatmap, and the parameter $\sigma$ represents GKS, which models the uncertainly of an annotation for a keypoint.

Due to the inconsistency between the size of the heatmap $H\in \mathbb{R}^{\frac{W}{R}\times \frac{H}{R}\times C}$ and the input image $I\in \mathbb{R}^{H \times W\times 3}$, where $H$ and $W$ are the height and width of an image respectively, $R$ ($R>1$) is the down-sampling rate, and $C$ is the number of keypoint classes, quantization errors inevitably occur during the process of the coordinate encoding in Eq\eqref{eqt1:gaussionGT}. For a GT keypoint $\bm{{\mu}} =[\mu_x,\mu_y]$ from the image, its corresponding coordinate in the heatmap is as follows:

\begin{equation}
\label{eqt3:round function}
\widetilde{\bm{{\mu}}}=\lfloor \frac{\bm{{\mu}}}{R} \rfloor
\end{equation}
where the operation $\lfloor\cdot \rfloor$ causes a quantization error:
\begin{equation}
\label{eqt3:error}
O = \frac{\bm{{\mu}}}{R}-\widetilde{\bm{{\mu}}}
\end{equation}
in which the range of the quantization error $O$ is bounded by $R\cdot[-1/2, 1/2)\times[-1/2, 1/2)$. As a result, the quantization error $O$ deeply influences the accuracy of the pixel-wise point location.

In this paper, we introduce a local offset in the position encoding process to reduce the quantization errors. Specifically, we use a deep model to predict the Gaussian heatmap $\hat H $ and the quantization error $\hat{O}$ as follows:
\begin{align}\label{eqt3:loss}
 L=\frac{1}{N}\sum_{k=1}^{N} \|H_i-\hat H_i\|^2 + \frac{\lambda}{N}\sum_{k=1}^{N} \|{O}_i-\hat{O}_i\|^2
 \end{align}
where $\lambda$ is a hyperparameter, $i$ ($1\leq i \leq N$) is the index of a sample. The loss function~\eqref{eqt3:loss} consists of two parts: one computes the mean square error between the GT heatmap $H$ and the predicted heatmap $\hat H$; the other computes the mean square error between the real quantization error ${O}$ and the predicted quantization error $\hat{O}$.

Once the Eq\eqref{eqt3:loss} is optimized, the predicted coordinate $(\hat x, \hat y)$ is decoded as follows:
\begin{equation}\label{eqt:decorder}
(\hat x, \hat y)= R\cdot\big(\text{Argmax}_{(x,y)}(\hat H) + \hat{O}\big)
\end{equation}
where the function $\text{Argmax}(\cdot)$ locates the spatial coordinate by finding the maximum value.

\begin{figure}[t!]
    \centerline{\includegraphics[scale=0.30]{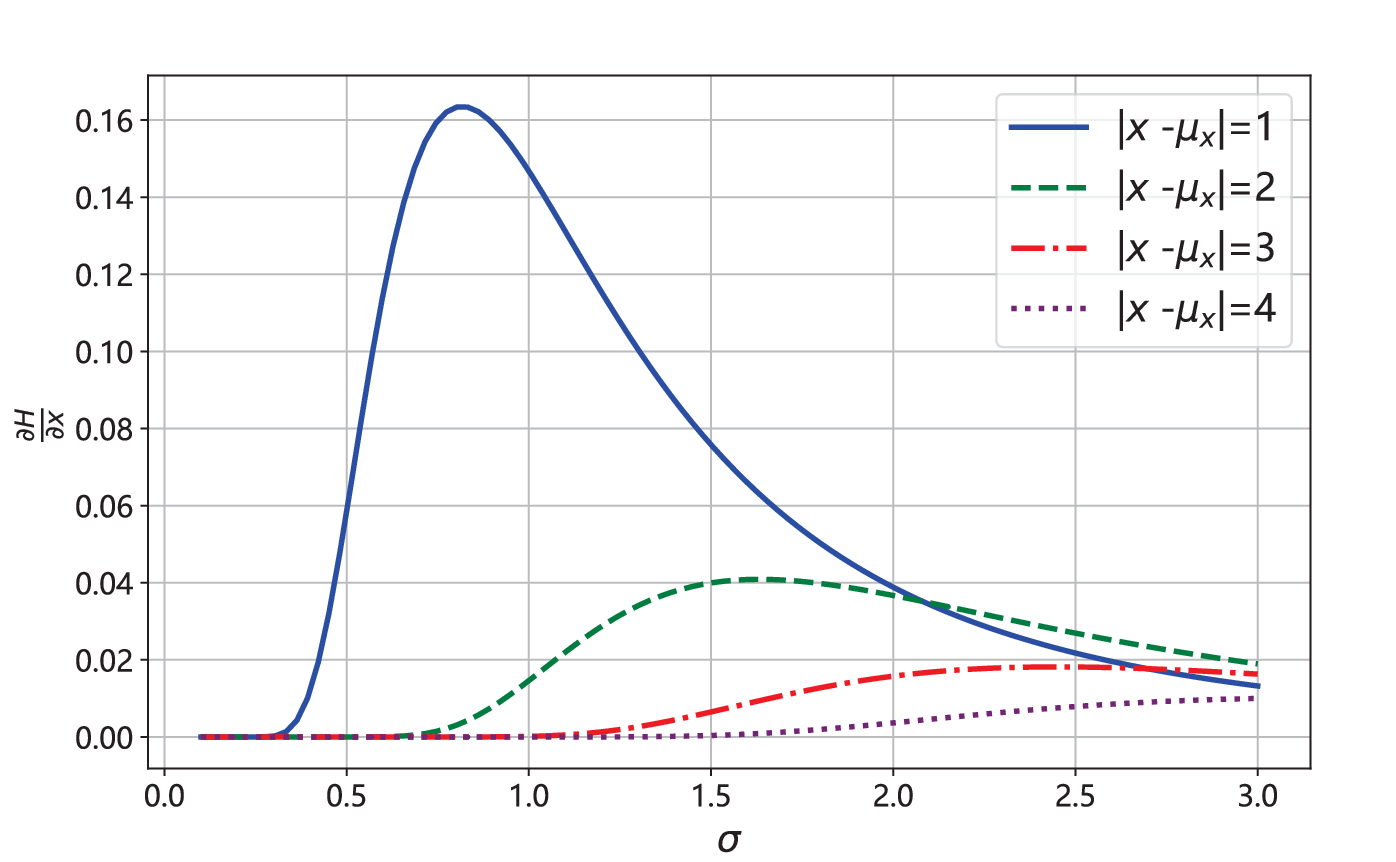}}
    \caption{The relationship between the gradients~\eqref{eqt4:gradient} and the distance $|x-\mu_x|$.}
    \label{fig-gradient}
\end{figure}

\subsection{Progressive Change of the GKS}\label{sec:progressive-change-of-GKS}
Heatmaps in Eq.\eqref{eqt1:gaussionGT} have a significant impact on the accuracy of coordinate decoding for each point. As illustrated in Fig.\ref{fig:combined_heatmap}, the larger GKS is, the smoother the heatmap distribution is. Consequently, during the training stage, a small disturbance would have a higher chance to change the decoded position for the larger GKS than the smaller one~\cite{nibali2018numerical}, although the larger GKS has a lower training loss in Eq\eqref{eqt3:loss} than for the smaller one. As illustrated in Fig.~\ref{fig:combined_heatmap}(c) and (d), for the pixel-wise location tasks, it is suitable to utilize a smaller GKS than a larger one due to the MVD problem.

On the other hand, a larger GKS would bring more supervised signal to train a deep model. Specially, the gradient of $H$ in Eq\eqref{eqt1:gaussionGT} with respect to each pixel coordinate is as follows:
\begin{equation}\label{eqt4:gradient}
{\frac{\partial H}{\partial x}}= {\frac{( {x - {\mu _x}})}{\sqrt {2\pi } {\sigma ^3}} }\exp \left( -\frac{(x - \mu_x)^2 + (y - \mu_y)^2}{2\sigma^2} \right)
\end{equation}
Fig.\ref{fig-gradient} illustrates the relationship between the a spatial point $x$ and a GT one $\mu_x$. We have drawn two key observations from Fig.\ref{fig-gradient} as follows: 1) For a fixed GKS, the farther a spatial point is away from the GT, the smaller the supervised signal is; 2) For a fixed spatial point, the larger a GKS is, the larger the supervised signal is.

\begin{figure}[t!]
    \centerline{\includegraphics[scale=0.35]{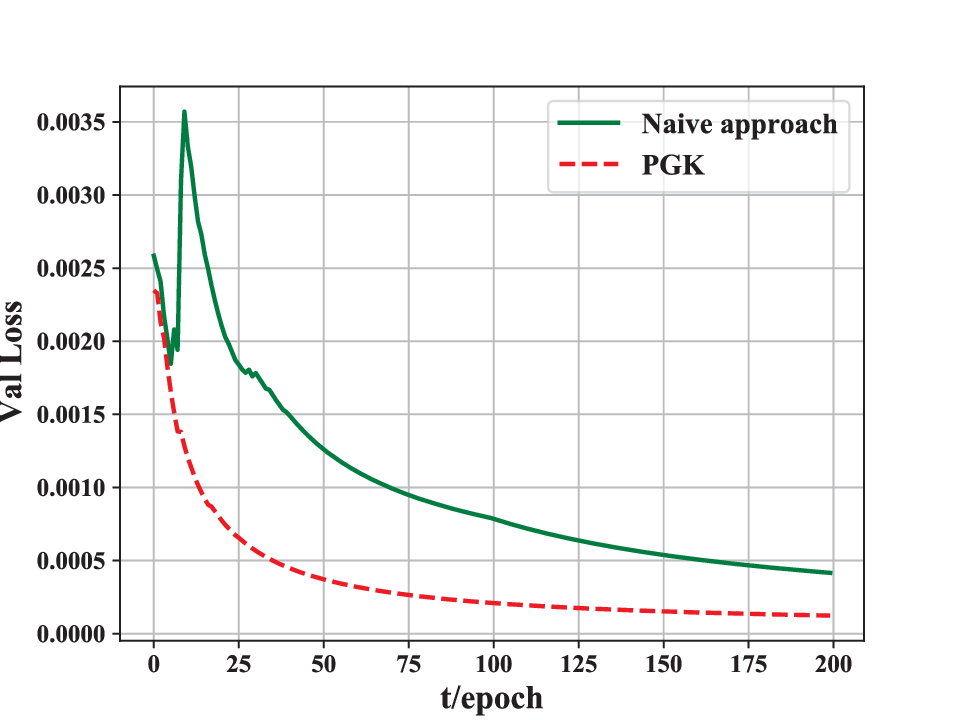}}
    \caption{The loss curve of the validation set for PGK and naive approach. At 100 epochs, the naive method changes the GKS from 3 to 1.}
    \label{val-loss}
\end{figure}

Therefore, a naive idea to utilize the above observations is to switch the GKS in the training of a model. That is, in the early stage of training, in order to obtain a large supervision signal for these points away from the GT ones, a larger GKS is used to help the model converge quickly; while, in the later stage of training, a smaller GKS makes the model focus on the local area around the GT points for the detailed information. However, the naive approach faces the ``switch problem": the weights from the larger GKS is not a good initialization for the smaller GKS. As illustrated in Fig.\ref{val-loss},  the naive approach did not accelerate the convergence of training process.

To handle the switch problem, we propose a progressive keypoint location
method referred to as point location via Progressive Gaussian Kernels in an annealing approach as follows:
\begin{equation}\label{eqt5:progressive-gaussian-Kernels-function}
{{\sigma }_{t}}=({{\sigma }_{max}}-{{\sigma }_{min}}) {{\alpha }^{t/M}}+{\sigma _{min}}{\big(1 - ({\sigma _{max}} - {\sigma _{min}}) \alpha \big) ^{t/M}}
\end{equation}
where the parameters $\sigma_{max}$ and $\sigma_{min}$ ($\sigma_{max}> \sigma_{min}$) represent the initial GKS and the target one, respectively. $\sigma_t$ is the annealing GKS at the $t$-th epoch, $\alpha$ ($0<\alpha<1$) is the annealing rate, and $M$ ($1\leq t\leq M$) is the total number of epochs during training.

\begin{figure}[t!]
    \centerline{\includegraphics[scale=0.35]{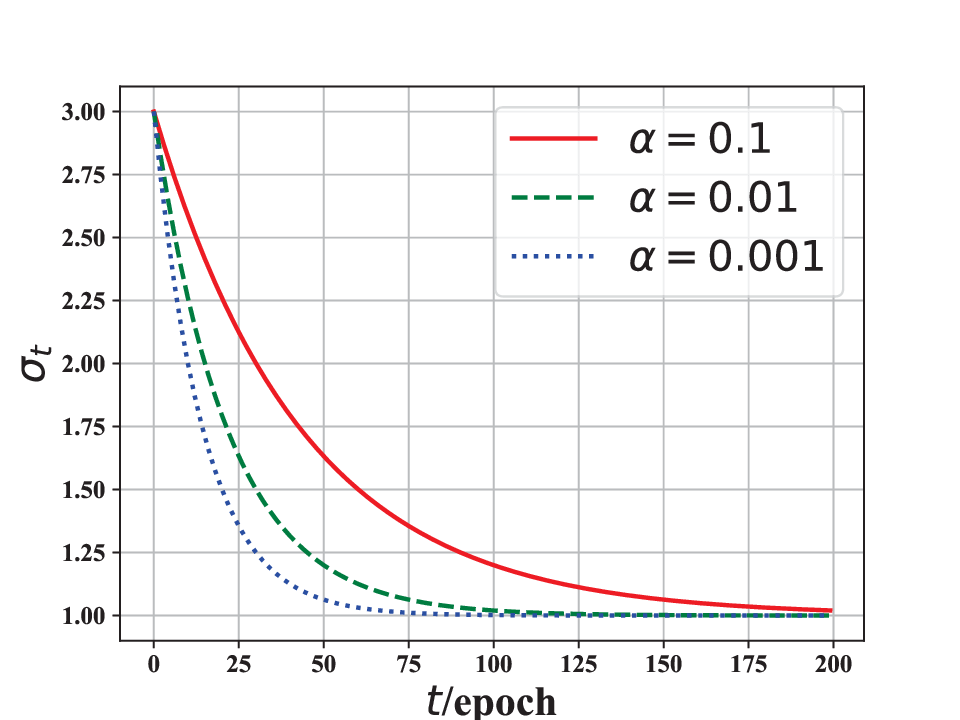}}
    \caption{The curves of GKS in Eq~\eqref{eqt5:progressive-gaussian-Kernels-function} with different values of $\alpha$, the other hyper-parameters are ${\sigma _{max}} = 3$, $\sigma _{min}=1$, and $M = {\rm{200}}$.}
    \label{fig4-progressive-Gaussian-kernel}
\end{figure}

The progressive Gaussian kernel is shown in Fig.\ref{fig4-progressive-Gaussian-kernel}. Following the idea of the annealing approach~\cite{barkhordari2021response}, the function ${\alpha ^{t/M}}$ progressively increases with respect to the number of the epochs. Specifically, the change rate of $\sigma_t$ is controlled by
 $({\sigma _{max}} - {\sigma _{min}}) {\alpha ^{t/M}}$, which reduces the GKS at each epoch.

\subsection{Grouping Keypoints into Rectangle Symbols}

The located keypoints have two types of errors: 1) the type of a keypoint is misclassified; 2) the location of a keypoint still has an error. We propose a strategy with the error-correction mechanism to group the keypoints into rectangle symbols for the classification error.

Concretely, we utilize the rectangle shape as a prior to design multiple grouping routes to remove classification errors according the consistence between the grouping point and the candidate one. We formally represent the consistence between points as: $1==\text{consistence}( p_i, p_j) $ . For example, point 7 and point 8, 12, 14 and 15 in Fig.~\ref{fig:keypoint-class-Diagram} have the consistent group direction to group two points into one third rectangle symbol if we group a rectangle from the right of point 7.
An example of grouping keypoints into rectangle symbols is illustrated in Alg.~\ref{alg:1}.

 \begin{algorithm}
    \renewcommand{\algorithmicrequire}{\textbf{Input:}}
    \renewcommand{\algorithmicensure}{\textbf{Output:}}
    \caption{Grouping keypoints into the predefined symbols}
    \label{alg:1}
    \begin{algorithmic}[1] 
        \REQUIRE A start point set $\mathcal{K}$ and the bounding box set of symbols $B_i \in \mathcal{B}$
        \ENSURE The rectangle symbol set $V_i \in \mathcal{V}$ where $|\mathcal{V}| = |\mathcal{B}|$
        \FOR{$B_j \in \mathcal{B}$}
            \STATE Select a start point $p_i \in \mathcal{K}$ as the grouping point from $B_j$
            \STATE  Follow the grouping direction of the point $p_i$ and calculate Euclidean distance $d_{ij}$ between the point $p_i$ and the candidate ones $p_j$
            \IF{$l=\min_j d_{ij}$ and 1==\text{consistence}( $p_i$, $p_l$)}
               \STATE Add the point $p_l$ into the matched list $V_i$
               \STATE Use the point $p_l$ as the new start point, $k_i\leftarrow p_l$
            \ELSE
                \STATE Retrieve anther keypoint from $V_i$ as new start point
            \ENDIF
            \STATE Goto Step~3
            \ENDFOR
            \STATE \textbf{return} $V_i$.
    \end{algorithmic}
\end{algorithm}

\section{Experiments}\label{sec:section4}

\subsection{Dataset for CAD images}

The dataset\footnote{https://github.com/pangjunbiao/CAD-dataset/} consists of 300 images for the training and 60 images for the testing, respectively. The image resolutions range from $1700\times 1200$ to $4200 \times 3000$. Some examples are illustrated in Fig.\ref{fig:cad-visualization-result}. We use Labelme as the annotation software. Because it supports various annotation methods, including point annotation, rectangle annotation, and polygon annotation. These CAD images about the layout of the equipment rooms are from the telecommunication industrial CAD drawings.

Fig.\ref{fig:group-meaningful-entities} shows 12 types of semantic symbols, including 3 types of rectangle symbols (\textit{i.e.}, Scales, Blocks, and Walls) and 9 types of region symbols (The types of symbols are illustrated in Table~\ref{tbl:obs-result}). The size of these symbols in CAD images are significantly different from each other.
The region symbols are efficiently located by object detection methods for the OBS task. Specifically, the sizes of the rectangle symbols are smaller than those of the region symbols, requiring the pixel-wise location.

\begin{figure}[t!]
\centerline{\includegraphics[scale=0.30]{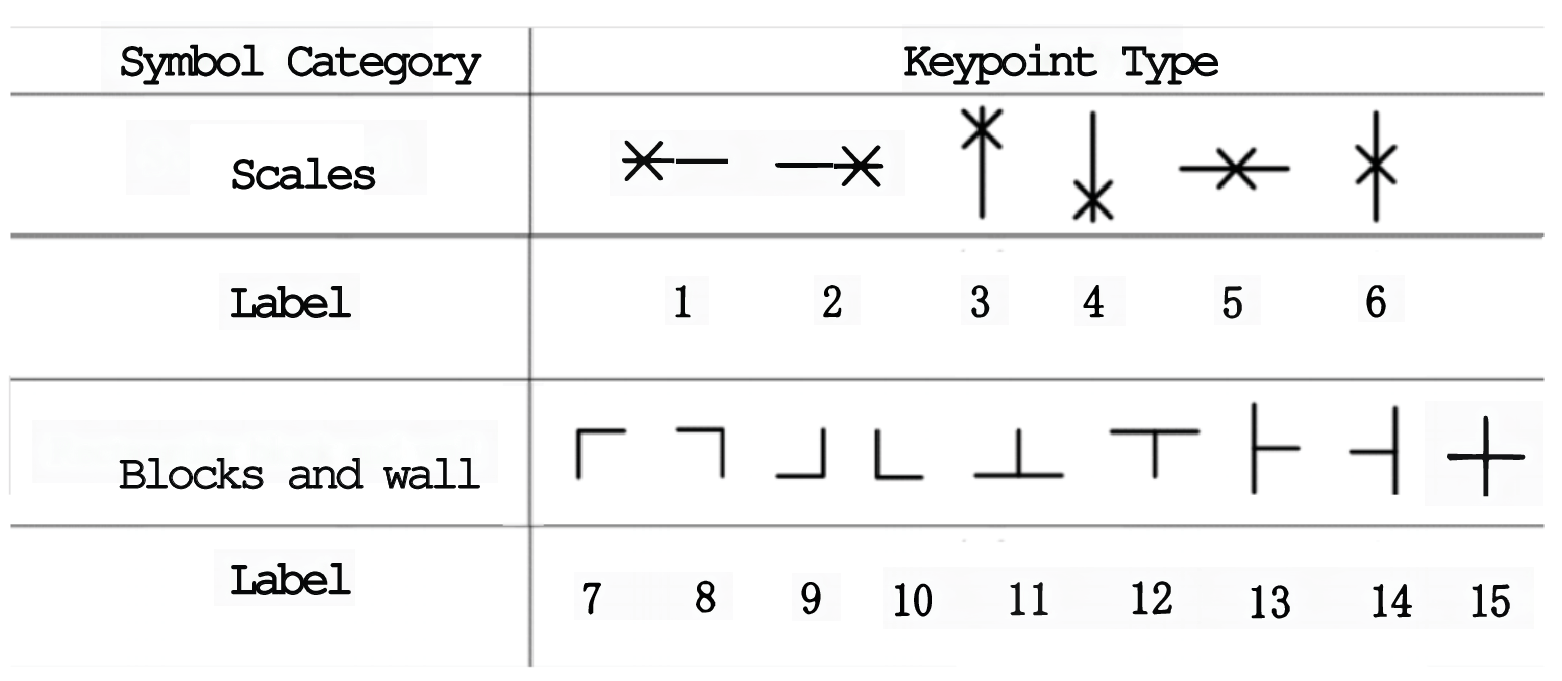}}
    \caption{15 kinds of keypoints are defined to group the rectangle symbols.}
    \label{fig:keypoint-class-Diagram}
\end{figure}

To locate the rectangle symbols, the first step involves coarsely localizing region symbols, followed by the point-wise keypoint location. Fig.~\ref{fig:keypoint-class-Diagram} shows the 15 types of keypoints that compose the rectangle symbols. Concretely, for scales, 6 types of keypoints are defined based on the direction of the scale's endpoints. For blocks and walls, 9 types of keypoints are defined based on the shape of the corners.

In summary, parsing rectangle symbols poses a significant challenge to the pixel-wise point location for the symbol spotting task. Because the symbol grouping algorithm in Alg.~\ref{alg:1} requires both the type and the location of keypoints should be correctly classified and located.

\subsection{Evaluation Metrics}

In our experiment, we used two kinds of evaluation metrics to evaluate the accuracy of the keypoint location and the symbols spotting as follows:

\textbf{F1 score:} The F1 score offers a balanced evaluation between Precision and Recall as follows:
\begin{align}\label{eqt:f1}
\text{F1} = \frac{{2 \times \text{Precision} \times \text{Recall}}}{{\text{Precision} + \text{Recall}}}
\end{align}
where Precision and Recall are respectively defined as follows:
\begin{align}\label{eqt:precision}
      \text{Precision} =\frac{\text{TP}}{\text{TP + FP}}
\end{align}
\begin{align}\label{eqt:recall}
      \text{Recall} =\frac{\text{TP}}{\text{TP + FN}}
\end{align}
in which TP is the number of the successfully detected keypoints or region symbols, FP is the number of incorrectly detected keypoints or region symbols, and FN is the number of keypoints or region symbols that were not detected when they should have been. Specifically, whether a keypoint or region symbol has been correctly detected is further defined as follows:
\begin{itemize}
    \item For region symbols, Intersection over Union (IoU) evaluates the ratio of the intersection and union between the predicted region symbols $B$ and true symbols $G$:
    \begin{equation}
        \text{IOU}=\frac{\text{Intersection}(B,G)}{\text{Union}(B,G)}
    \end{equation}
When $\text{IOU}$ is larger than a predefined threshold $\tau_o$( i.e., $\text{IOU} > \tau_o$), we consider that a region symbol is correctly detected. In this paper,  $\tau_o$ is set as $0.5$.
\item For keypoints, Euclidean distance $\ell_i$ between the $i$-th predicted keypoint and the corresponding GT point is used:
\begin{equation}\label{eqt:ell-between-points}
    \ell_i= \sqrt{(\hat{x}_i-x_i)^2+(\hat{y}_i-y_i)^2}
\end{equation}
When $\ell_i$ is smaller than a predefined threshold $\tau_p$ ( i.e., $\ell_i < \tau_p$), we consider that a keypoint is correctly detected. In this paper, $\tau_p$ is set as 2.
\end{itemize}

 \textbf{Averaged Pixel Error for Keypoints (APEK):} APEK evaluates the accuracy of the pixel-wise keypoint location as follows:
\begin{align}\label{eqt:apek}
\text{APEK} = \frac{1}{N}\sum\limits_{i=1}^N {{\ell_i}}
\end{align}
where $\ell_i$ is the Euclidean distance in Eq\eqref{eqt:ell-between-points}, $N$ is the number of all the detected keypoints (including the wrongly detected ones).

\subsection{Implement Details}

We use the Stacked Hourglass network~\cite{newell2016stacked} as the backbone network.
The initial learning rate for the ADAM optimizer was set to $10^{-4}$, and the hyper-parameter ${\lambda}$ was set to {\rm{0}}{\rm{.1}}. The model was optimized for $200$ epochs.
In the training stage, we first locate the region symbols, the detected region symbols are cropped into a set of the $256\times 256$ pixel patches by a sliding window approach with the step size is equal to 0.
In the inference stage, we divide a complete CAD image into pixel patches of size $256\times 256$, and then feed each patch into the model for keypoint localization, where Non-Maximum Suppression (NMS) is applied to obtain the keypoints. We detected a keypoint by judging whether the max value of the heatmap $\hat H$ in Eq\eqref{eqt:decorder} is larger than the threshold $\varsigma  = 0.6$. Finally, we stitch the results from each patches into the coordinates of the CAD images.
\subsection{Experimental Results}

The rectangle symbols usually are contained within these region symbols. Rather than detecting rectangle symbols across the whole CAD images, it is reasonable to locate the rectangle symbols within the region symbols.

\begin{table}[t!]
\centering
\caption{The results of the object-based symbol spotting by object detection method}
\label{tbl:obs-result}
\begin{tabular}{|c|c|c|c|}
\hline
Region Symbol & Precision & Recall & F1\\
\hline\hline
Main & 1.00 & 1.00 & 1.00\\ \hline
Main title & 1.00 & 0.92 & 0.96\\ \hline
Title bar & 1.00 & 1.00 & 1.00\\ \hline
Material list & 1.00 & 1.00& 1.00 \\ \hline
Legend  & 1.00 & 1.0& 1.00\\ \hline
Description text & 1.00 & 1.00 & 1.00\\ \hline
Direction angle & 1.00 & 0.90 &0.95\\ \hline
Text with directional lines & 1.00 & 0.98 & 0.99\\ \hline
Scale text & 0.97 & 0.97 & 0.97 \\ \hline
\end{tabular}
\end{table}
\subsubsection{Performances of OBS}\label{sec:Performance of OBS}

 We located the region symbol by YOLOX~\cite{ge2021yolox}. Table~\ref{tbl:obs-result} demonstrates that YOLOX achieves an excellent performance for the object-based symbol spotting. Serving as the layout analysis for CAD images, the accuracy of the OBS task is critical to improve the inference speed for the PBS task. Because rather than locating the small rectangle symbols in a high-resolution image (see Fig.\ref{fig:cad-visualization-result}), we need to further finely locate the rectangle symbols within these detected region symbols.

 In this work, the comparisons with these advanced methods (such as~\cite{wang2023yolov7}~\cite{reis2023real}) are not conducted. Because the OBS task is not our main contribution in this work; besides, we believe that any advanced object detection methods could be used to the OBS task.

\subsubsection{Performances of PBS}

\textbf{Baselines} There are many heatmap-based methods and the regression-based methods for point location for HPE. To show the superiority of the pixel-wise location ability of the proposed method for parsing CAD drawings, we choose two State-Of-The-Art (SOTA) methods (\textit{i.e.}, soft-argmax~\cite{sun2018integral}) from the former and one (\textit{i.e.}, RLE~\cite{li2021human}) from the latter as our baselines as follows:
\begin{itemize}
    \item \textbf{Soft-argmax~\cite{li2021localization}} assumes that a model learns a discrete probability map $\pi_{y_i}$ which indicates the probability of the predicted target point at $y_i$. The quantization error is avoided by an elegant approximation, \textit{i.e.}, $\hat{y} = \text{soft-argmax}(\bm{\pi})= \sum_i\pi_{y_i}y_i$, where $\bm{\pi}$ is a normalized distribution. Soft-argmax avoids the quantization errors by the expectation of the probability map $\bm{\pi}$.
    \item \textbf{RLE~\cite{li2021human}:} RLE utilizes the normalizing flows~\cite{rezende2015variational} to capture the underlying output distribution and makes the regression-based methods match the accuracy of SOTA heatmap-based methods. RLE, as a regression-based method, naturally avoids the quantization errors.
\end{itemize}

To achieve the pixel-based symbol spotting, we firstly locate keypoints in Fig.\ref{fig:keypoint-class-Diagram} via PGK in subsection~\ref{sec:progressive-change-of-GKS}, and then apply Alg.~\ref{alg:1} to group these detected keypoints into rectangle symbols. In the following experiments, we report the performances of keypoint location and then the accuracy of the Alg.~\ref{alg:1}.

\textbf{Performances of Keypoint Location.} Table~\ref{tbl:symbol-spotting-results} shows the consistent performance superiority over the heatmap-based counterpart, \textit{i.e.}, soft-argmax~\cite{sun2018integral} and the regression-based method, \textit{i.e.}, RLE~\cite{li2021human}. Note that the evaluation and network training are conducted under the same input size. Table~\ref{tbl:symbol-spotting-results} also shows the offset component in Eq\eqref{eqt3:loss} uniformly increase the performances keypoint location in terms of both F1 and AEPK measurements. For example, our method outperforms heatmap-based counterpart by +0.03 F1 and +0.41 AEPK, respectively.

\textbf{Performances of Symbol Grouping Algorithm.}

\begin{table}[t!]
\centering
\caption{The results of point location and the symbol grouping algorithm}
\begin{tabular}{|c|c|c||c|c|}
\hline
\multirow{2}{*}{Methods} & \multicolumn{2}{c||}{Point} & Scale & Block and Wall \\ \cline{2-5}
 & F1 $\uparrow$ & AEPK $\downarrow$  & \multicolumn{2}{c|}{F1 $\uparrow$} \\ \hline\hline
soft-argmax& 0.76 & 4.92  & 0.77 & 0.76  \\ \hline
RLE& 0.78 & 4.85 & 0.79 & 0.78  \\ \hline
Our method &\textbf{ 0.79} & \textbf{4.51} & \textbf{0.79} & \textbf{0.79}  \\ \hline
\end{tabular}
\label{tbl:symbol-spotting-results}
\end{table}

Table~\ref{tbl:symbol-spotting-results} also shows consistent performance gains over the heatmap-based counterparts, \textit{i.e.}, soft-argmax~\cite{sun2018integral} and the regression-based method, \textit{i.e.}, RLE~\cite{li2021human}. Results presented in Table~\ref{tbl:symbol-spotting-results} demonstrate that:1) the accuracy of point location has a great influence on the final grouped symbols; 2) the symbol grouping alg.~\ref{alg:1} is able to partially rectify some misclassified points.

In summary, according to the results presented in Table~\ref{tbl:symbol-spotting-results}, we can further draw the following conclusions: 1) Soft-argmax still produces an unconstrained probability map for CAD images, which also suffers the bias problem~\cite{gu2023bias}.
2) Our proposed method combines the characteristics of the heatmap-based method and the regression-based one, leading to a simpler and more efficient scheme.

\subsubsection{Ablation Studies for KeyPoint Location}

\begin{table*}[t!]
\centering
\caption{The effectiveness of the local offset for keypoint location}
\scalebox{0.8}{
\begin{tabular}{|c|c|c|c|c|c|c|c||c|c|c|c|c|c|c|c|c|c|c|c|}
\hline
\multirow{2}{*}{}& \multirow{2}{*}{local offset} & \multicolumn{6}{c||}{Scale}   & \multicolumn{9}{c|}{Block and Wall}&
\multirow{2}{*}{Average}\\  \cline{3-17}
&  &point1&point2&point3&point4&point5&point6&point7&point8&point9&point10&point11&point12&point13&point14&point15&\\ \cline{2-18}\hline\hline
\multirow{2}{*}{F1$\uparrow$}
&\ding{55}&0.70&0.70&0.72&0.83&0.78&0.84&0.79&0.81&0.75&0.80&0.83&0.78&0.90&0.77&0.70&0.77\\
\cline{2-18}
&$\checkmark$&\textbf{0.72}&\textbf{0.71}&\textbf{0.74}&\textbf{0.85}&\textbf{0.79}&\textbf{0.85}&\textbf{0.80}
&\textbf{0.81}&\textbf{0.76}&\textbf{0.82}&\textbf{0.86}&\textbf{0.81}&\textbf{0.92}&\textbf{0.80}&\textbf{0.72}& \textbf{0.79}\\
\cline{2-18}\hline
\multirow{2}{*}{AEPK$\downarrow$}
 &\ding{55}&5.00&4.88&4.99&4.76&4.9&4.51&4.59&4.94&4.74&4.89&4.73&5&4.84&4.91&4.92 &4.88\\ \cline{2-18}
 &$\checkmark$&\textbf{4.78}&\textbf{4.58}&\textbf{4.56}&\textbf{4.19}&\textbf{4.34}&\textbf{4.23}&\textbf{4.5}
 &\textbf{4.47}&\textbf{4.34}&\textbf{4.72}&\textbf{4.44}&\textbf{4.73}&\textbf{4.51}&\textbf{4.69}&\textbf{4.77} &\textbf{4.51}\\ \cline{2-18} \hline
\end{tabular}}
\label{tbl:Effectiveness-of-Local-Offset}
\end{table*}

\begin{table*}[t!]
\centering
\caption{The effectiveness of PGK for keypoint location}
\scalebox{0.8}{
\begin{tabular}{|c|c|c|c|c|c|c|c||c|c|c|c|c|c|c|c|c|c|c|c|}
\hline
\multirow{2}{*}{}& \multirow{2}{*}{PGK}  & \multicolumn{6}{c||}{Scale}   & \multicolumn{9}{c|}{Block and Wall}&\multirow{2}{*}{Average}\\  \cline{3-17} &&point1&point2&point3&point4&point5&point6&point7&point8&point9&point10&point11&point12&point13&poin14&point15&\\ \cline{2-18}\hline\hline
\multirow{2}{*}{F1 $\uparrow$}
&\ding{55}
&0.72&0.71&0.74&0.85&0.79&0.85&0.80&0.81&0.76&0.82&0.86&0.81&0.92&0.80&0.72& 0.79\\ \cline{2-18}
&$\checkmark$&\textbf{0.75}&\textbf{0.79}&\textbf{0.81}&\textbf{0.9}&\textbf{0.87}&\textbf{0.89}
&\textbf{0.83}&\textbf{0.87}&\textbf{0.77}&\textbf{0.83}&\textbf{0.90}&\textbf{0.86}&\textbf{0.95}&\textbf{0.83}&\textbf{0.78}&\textbf{0.84}
\\ \cline{2-18}\hline
\multirow{2}{*}{AEPK$\downarrow$}
 &\ding{55}&4.78&4.58&4.56&4.19&4.34&4.23&4.5&4.47&4.34&4.72&4.44&4.73&4.51&4.69&4.77 &4.51\\ \cline{2-18}
 &$\checkmark$&\textbf{4.62}&\textbf{4.56}&\textbf{4.41}&\textbf{3.95}&\textbf{4.33}&\textbf{4.18}
 &\textbf{4.33}&\textbf{4.29}&\textbf{4.08}&4.51&\textbf{4.43}&\textbf{4.43}&\textbf{4.26}&\textbf{4.48}&\textbf{4.52}&\textbf{4.36}
 \\ \cline{2-18} \hline
\end{tabular}}
\label{tbl:Effectiveness-of-PGK}
\end{table*}

In order to demonstrate the effectiveness of the local offset in Eq\eqref{eqt3:loss} and the PGK in Eq\eqref{eqt5:progressive-gaussian-Kernels-function}, we conducted ablation studies respectively.

\textbf{Effectiveness of Local Offset:} Table~\ref{tbl:Effectiveness-of-Local-Offset} shows that the local offset boosts performances of different types of keypoints in terms of both F1 and AEPK metrics.
Specifically, the performances of point 3 and point 4 in Fig.~\ref{fig:keypoint-class-Diagram} from the scale symbol are significantly improved from 0.72 to 0.74 and 0.83 to 0.85 respectively in terms of F1 score. Interestingly, The AEPK scores of the point 3 and point 4 are also reduced from 4.99 to 4.56 and 4.76 to 4.19, respectively. The results mean that the offset term not only improves the classification accuracy of points but also reduces the spatial location errors.

\begin{figure}[h!]
    \centering
    \centerline{
        \includegraphics[scale=0.40]{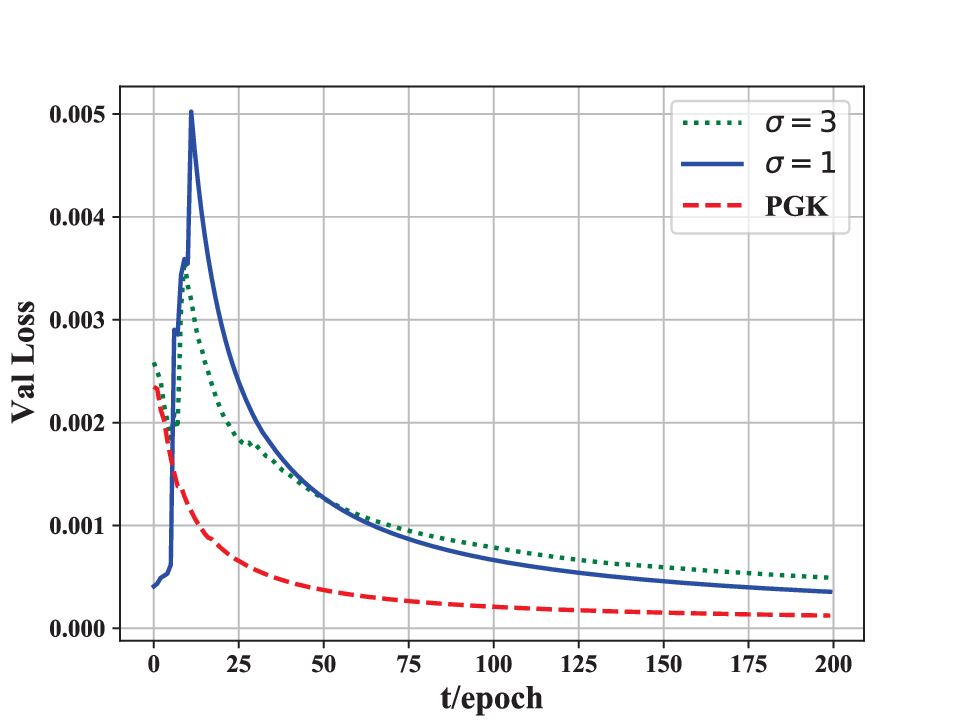}
    }\caption{Validation set loss curve with different Gaussian kernel.}
    \label{fig:val-speed}
\end{figure}
\begin{figure*}[t!]
    \centering
    \subfloat[$\sigma = 3$]  
    {
        \includegraphics[scale=0.5]{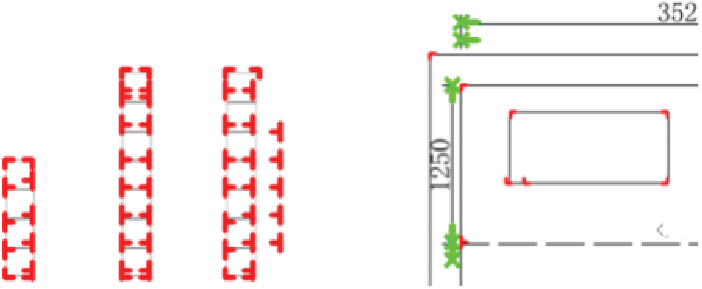}
        \label{fig:vision gks=3}
    }
    \subfloat[$\sigma = 1$]  
    {
        \includegraphics[scale=0.5]{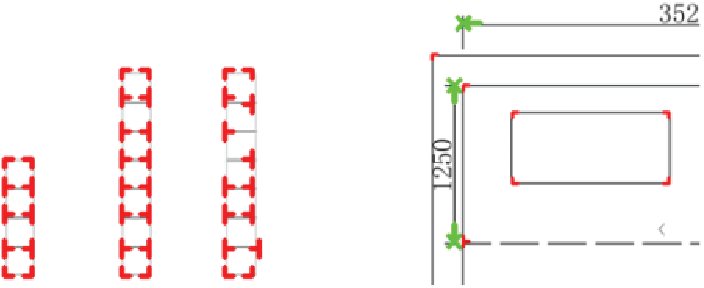}
        \label{fig:vision gks=1}
    }
    \subfloat[PGK]  
    {
        \includegraphics[scale=0.2]{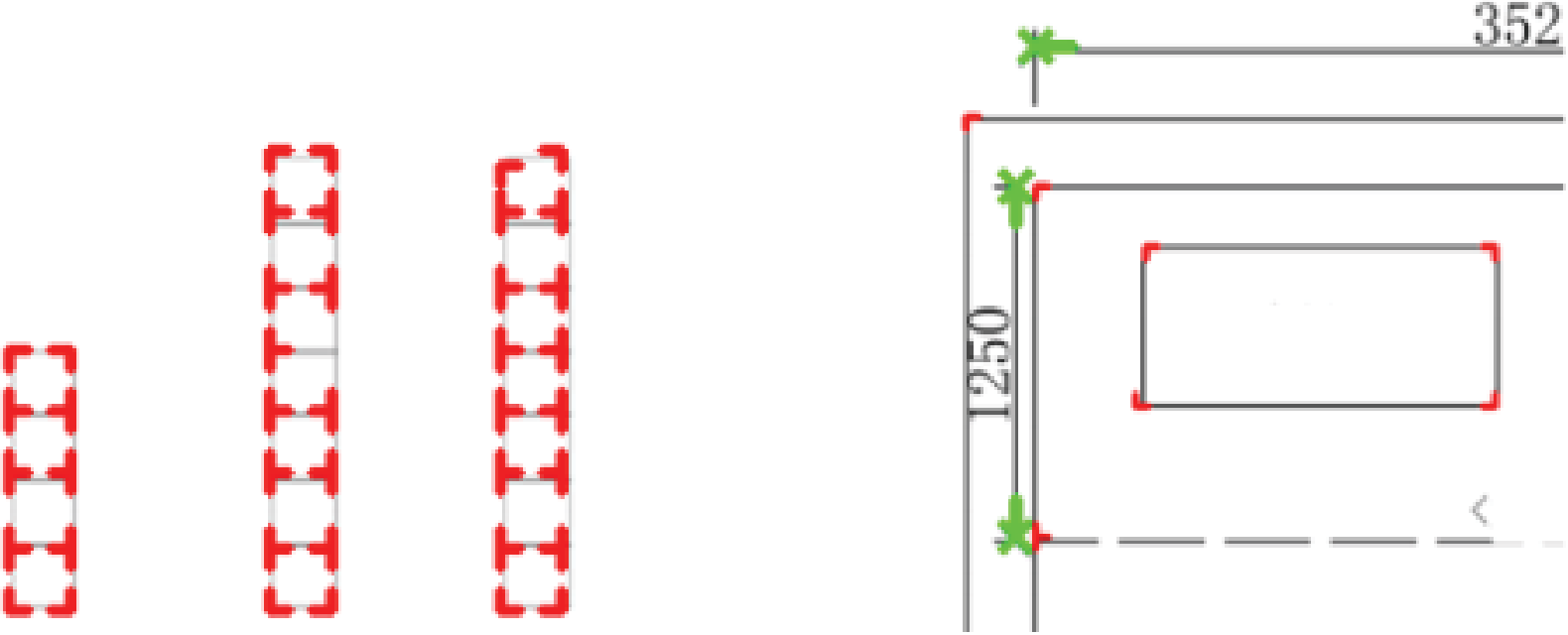}
        \label{fig:vision pgk}
    }
    \caption{Visualization results of the detected keypoint for the different $\sigma$.}
    \label{fig:Visualization-results-of-keypoint}
\end{figure*}

\begin{figure*}[t!]
\begin{center}
    \centering
    \centerline
    \subfloat

    {
        \includegraphics[scale=0.35]{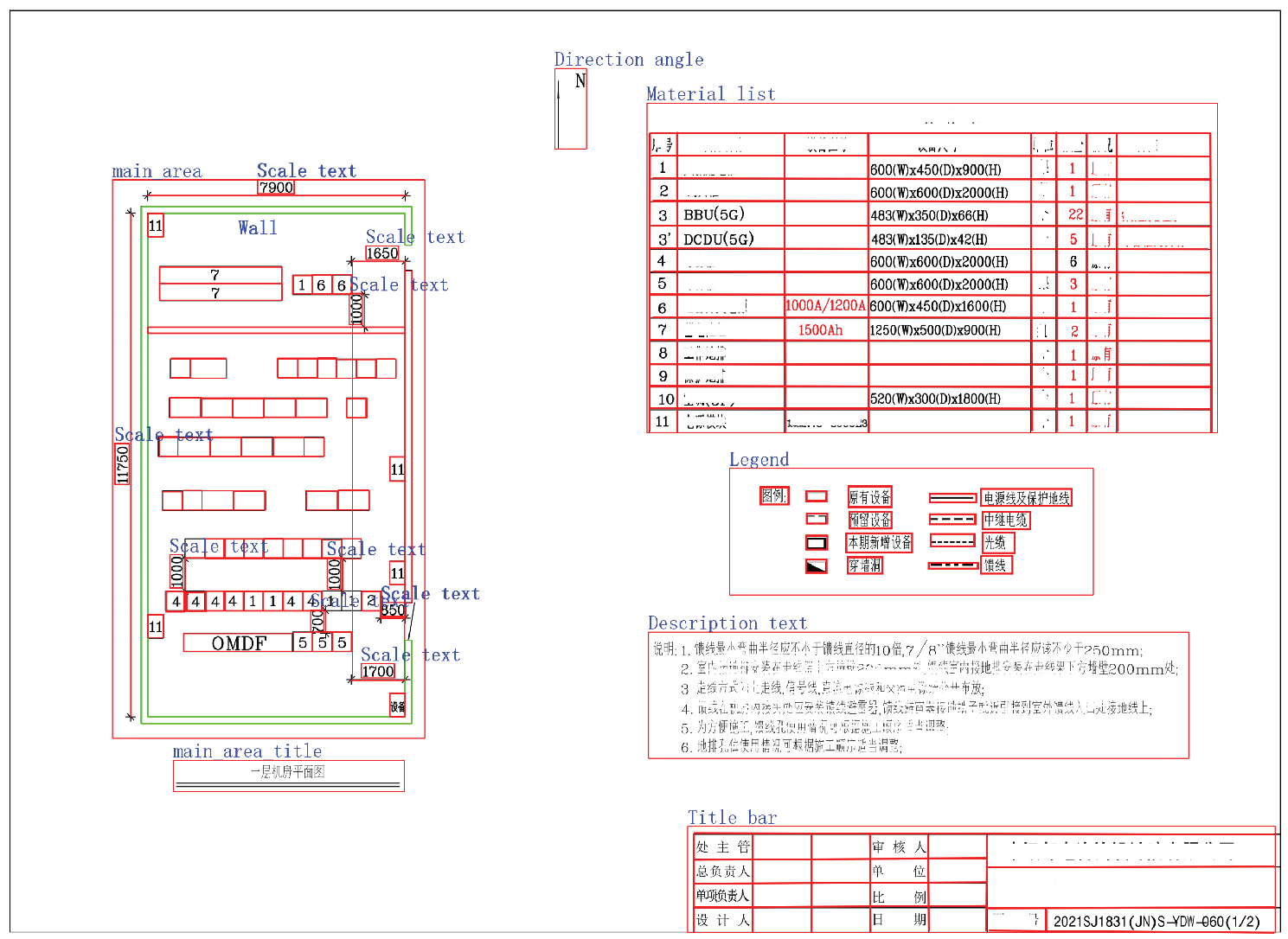}
    }
        \subfloat
    {
        \includegraphics[scale=0.43]{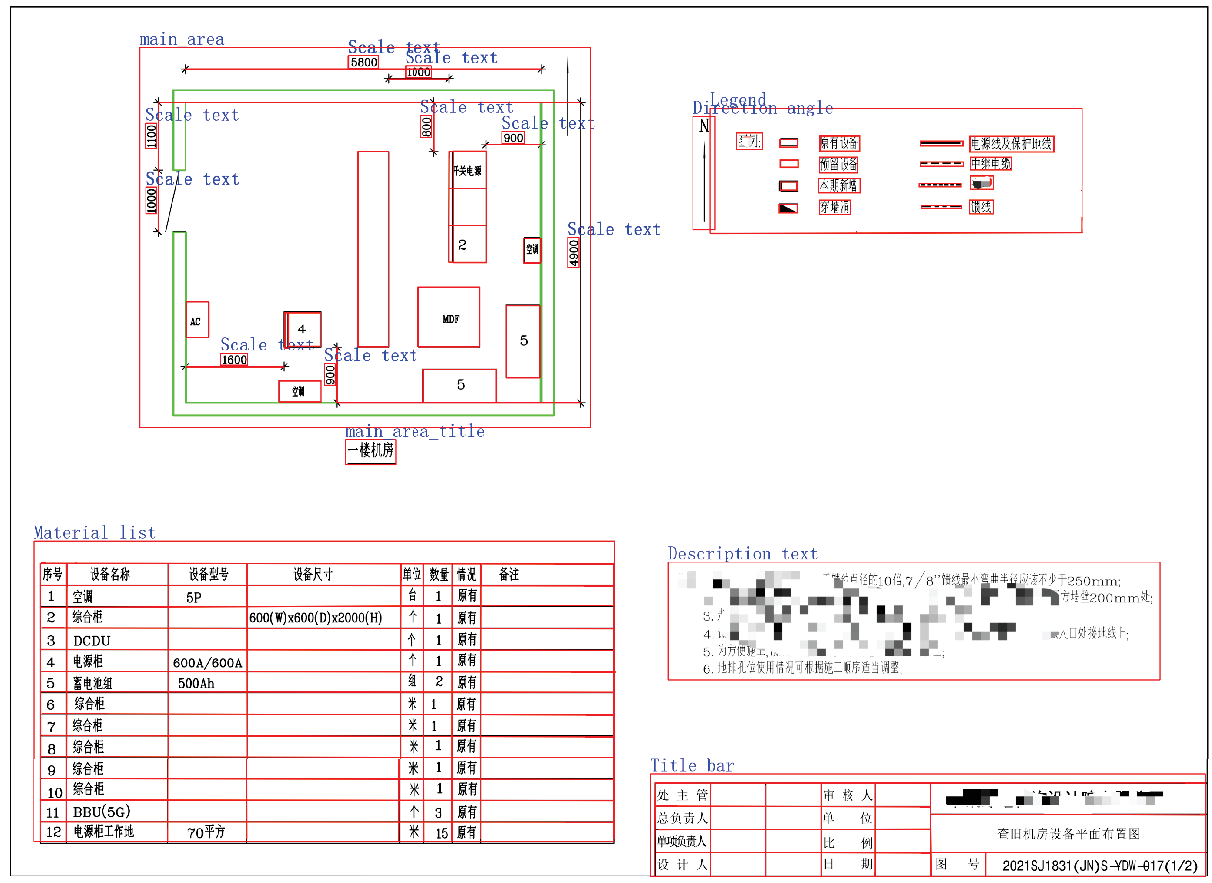}
    }
    \caption{sample results of the symbol spotting for CAD images.}
    \label{fig:cad-visualization-result}
\end{center}
\end{figure*}

\textbf{Effectiveness of PGK:} We discuss the impact of the proposed PGK by comparing the following baselines:
\begin{itemize}
    \item [1.] GKS with $\sigma=3$: $\sigma=3$ is a common setting to handle the uncertainty in the label errors for HPE. However, GKS with $\sigma=3$ would incur location errors due to the MVD problem in Fig.\ref{fig:combined_heatmap}.
    \item [2.] GKS with $\sigma=1$: $\sigma=1$ is barely used for HPE. We verify that a smaller GKS would obtain a lower location error than that of GKS with a larger kernel for symbol spotting from CAD images.
\end{itemize}
Fig.\ref{fig:val-speed} shows that the proposed PGK enables a model to not only have a ``smooth'' training process but also a faster convergence speed than the counterparts, \textit{i.e.}, GKS with $\sigma=3$ and GKS with $\sigma=1$.
 The comparison shows that the PGK enjoys the advantages of both the large GKS and the small one. Moreover, the smallest validation loss indicates that the proposed PGK has a best generalization ability compared to these counterpart methods. Fig.\ref{fig:Visualization-results-of-keypoint} vividly compares the detection results among different kernel sizes. Fig.\ref{fig:Visualization-results-of-keypoint}(a) shows that the GKS with $\sigma=3$ tends to predict duplicated keypoints due to the MVD problem and the NMS operation. Comparing between Fig.\ref{fig:Visualization-results-of-keypoint}(b) and Fig.\ref{fig:Visualization-results-of-keypoint}(c), the proposed PGK yields more accurate results than that of GKS with $\sigma=1$.

Results presented in Table~\ref{tbl:Effectiveness-of-PGK} further demonstrate that the proposed PGK shows consistent performance superiority over GKS with $\sigma=1$. For example, F1 score is increased from $0.79$ to $0.84$, and the averaged value of AEPK is decreased from $4.51$ to $4.36$.

\subsection{Visualization results}

Fig.\ref{fig:cad-visualization-result} illustrates that the visual result of the pixel-wise symbol spotting via both PGK and the symbol grouping algorithm. By comparing the spotting position with the original points, our system successfully parses the complex symbols for the telecommunication industrial drawings.

\section*{Conclusion}

In this paper, we propose the PGK approach to achieve both fast training speed and the good results in keypoint location. It adjusts the GKS in a progressively annealing approach to avoid the inefficiency of the small GKS yet enjoys the fast training speed of a large GKS. To enhance the location precision, we introduce a local offset into the heatmap based method. This approach enhances the accuracy of point location, setting a new benchmark in pixel-wise keypoint detection for CAD image parsing. We further propose a symbol grouping method with the error-correction ability, which is a significant advancement in the field of symbol spotting in CAD images.

Our method is computationally simple, coupled with its outstanding performance, underscoring its potential as a transformative tool for professionals engaged in CAD image analysis.By setting a precedent in the use of PGK and localized position encoding, this work paves the way for future investigations into efficient and accurate symbol spotting methodologies for parsing CAD images.

In the future work, we aim to enhance our work as follows: 1) how to group the non-rectangle symbols, such as, the symbols grouped by arcs as illustrated in Fig.~\ref{fig2-compare-cad-image-and-annotation}; 2) verify advantageous backbones, such as, HRNet~\cite{yu2021lite} and TokenPose~\cite{li2021tokenpose}, in terms of the training speed, inference speed and accuracy.

\bibliographystyle{IEEEtran}
\bibliography{ref}

\end{document}